\documentclass[10pt,twocolumn,letterpaper]{article}

\usepackage[pagenumbers]{cvpr} 

\usepackage[numbers]{natbib}
\usepackage{amsmath,amssymb}

\usepackage{multirow}

\usepackage[dvipsnames]{xcolor}

\definecolor{cvprblue}{rgb}{0.21,0.49,0.74}
\usepackage[pagebackref,breaklinks,colorlinks,citecolor=cvprblue]{hyperref}

\def\paperID{6361} 
\def\confName{CVPR}
\def\confYear{2024}

\newcommand{\ARCH}{SceneTex\xspace}

\newcommand{\mypara}[1]{\noindent\textbf{#1}}

\title{SceneTex: High-Quality Texture Synthesis for Indoor Scenes via Diffusion Priors}

\author{
Dave Zhenyu Chen$^{1}$ \quad Haoxuan Li$^{1}$ \quad Hsin-Ying Lee$^{2}$ \quad Sergey Tulyakov$^{2}$ \quad Matthias Nie{\ss}ner$^{1}$\\
$^{1}$Technical University of Munich \qquad $^{2}$Snap Research \\
\url{https://daveredrum.github.io/SceneTex/} \\
}

\begin{document}


\twocolumn[{%
	\renewcommand\twocolumn[1][]{#1}%
	\maketitle
	\thispagestyle{empty}
	\begin{center}
		\includegraphics[width=\textwidth]{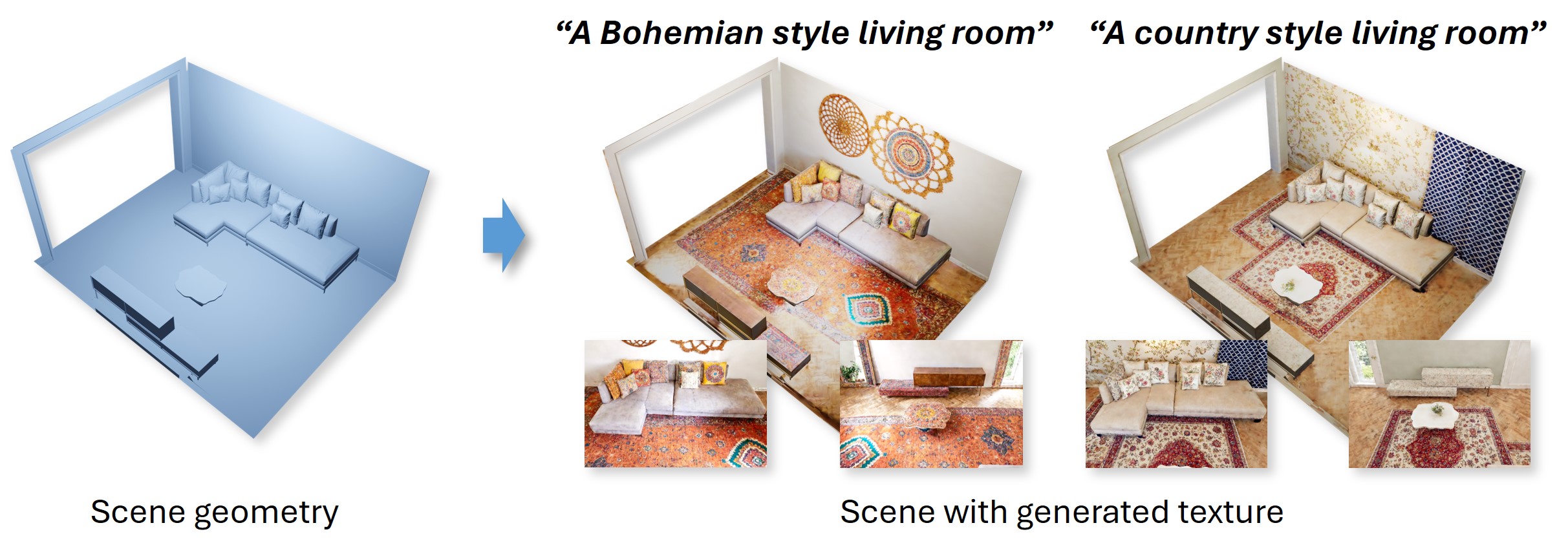}
		\captionof{figure}{
		We introduce~\ARCH, a text-driven texture synthesis architecture for 3D indoor scenes. Given scene geometries and text prompts as input, ~\ARCH generates high-quality and style-consistent textures via depth-to-image diffusion priors.
		}
		\label{fig:teaser}
	\end{center}
}]

\begin{abstract}

We propose SceneTex, a novel method for effectively generating high-quality and style-consistent textures for indoor scenes using depth-to-image diffusion priors. 
Unlike previous methods that either iteratively warp 2D views onto a mesh surface or distillate diffusion latent features without accurate geometric and style cues, SceneTex formulates the texture synthesis task as an optimization problem in the RGB space where style and geometry consistency are properly reflected. 
At its core, SceneTex proposes a multiresolution texture field to implicitly encode the mesh appearance. 
We optimize the target texture via a score-distillation-based objective function in respective RGB renderings. 
To further secure the style consistency across views, we introduce a cross-attention decoder to predict the RGB values by cross-attending to the pre-sampled reference locations in each instance.
SceneTex enables various and accurate texture synthesis for 3D-FRONT scenes, demonstrating significant improvements in visual quality and prompt fidelity over the prior texture generation methods.

\end{abstract}     
\section{Introduction}

Synthesizing high-quality 3D contents is an essential yet highly demanding task for numerous applications, including gaming, film making, robotic simulation, autonomous driving, and upcoming VR/AR scenarios.
With an increasing number of 3D content datasets, the computer vision and graphics community has witnessed a soaring research interest in the field of 3D geometry generation~\cite{smith2017improved, xie2018learning, lin2023infinicity, achlioptas2018learning, luo2021diffusion, zhang2021sketch2model, chen2019learning, autosdf2022}.
Despite achieving a remarkable success in 3D geometry modeling, generating the object appearance, i.e. textures, is still bottlenecked by laborious human efforts.
It typically requires a substantially long time for designing and adjustment, and immense 3D modelling expertise with tools such as Blender.
As such, automatic designing and augmenting the textures has not yet been fully industrialized due to a huge demand for human expertise and financial expenses.

Leveraging the recent advances of 2D diffusion models, tremendous progress has been made for text-to-3D generation, especially for synthesizing textures of given shapes~\cite{chen2023text2tex, richardson2023texture, metzer2022latent}. 
Seminal work such as Text2Tex~\cite{chen2023text2tex} and Latent-Paint~\cite{metzer2022latent} have achieved great success in generating high-quality appearances for objects, facilitating high-fidelity texture synthesis from input prompts.
Despite the fascinating results on objects, upscaling these methods to generating textures for an entire scene still confronts several challenges.
On one hand, methods that autoregressively project 2D views to 3D object surface~\cite{richardson2023texture, chen2023text2tex} usually suffer from texture seams, accumulated artifacts, and loop closure issues.
It is also quite difficult to maintain style consistency in the scene if every object is textured individually.
On the other hand, score-distillation-based approaches~\cite{metzer2022latent} perform texture optimization in the low-resolution latent space, often resulting in blurry RGB textures and incorrect geometry details.
As such, previous text-driven attempts fail to deliver high-quality textures for 3D scenes.

To address the aforementioned challenges, we propose SceneTex, a novel architecture to generate high-quality and style-consistent texture for indoor scene meshes by leveraging depth-to-image diffusion priors. 
Unlike previous methodologies that iteratively warp 2D views onto mesh surfaces, we take a different approach by framing the texture synthesis as a texture optimization task in RGB space via diffusion priors.
At its core, we introduce a multiresolution texture field to implicitly represent the appearance of the mesh.
To faithfully represent the texture details, we adopt a multiresolution texture to store texture features at multiple scales.
This enables our architecture to flexibly learn both low and high frequency appearance information.
To secure the style consistency of the generated texture, we incorporate a cross-attention decoder to reduce style incoherence introduced by self-occlusion.
Concretely, each decoded RGB values are produced by cross-attending to the pre-sampled reference surface locations scattered across each object.
This way, we further secure the global style consistency within each instance, as every visible location receives a global reference to the whole instance appearance.

We show that SceneTex has the capacity to facilitate versatile and accurate texture synthesis for indoor scenes with given language cues. 
We demonstrate in extensive experiments that SceneTex places a strong emphasis on both style and geometry consistency. 
The proposed method performs favorably against other text-driven texture synthesis methods in terms of 2D metrics such as CLIP score~\cite{radford2021learning} and Inception Score~\cite{smith2017improved}, and user study on a subset of the 3D-FRONT dataset~\cite{fu20213dfront}. 

We summarize our technical contributions as follows:
\begin{itemize}

\item We design a novel framework for generating high-quality scene textures in high resolution using depth-to-image diffusion priors. 

\item We propose an implicit texture field to encode the object appearance at multiple scales, leveraging a multiresolution texture to faithfully represent rich texture details.

\item We incorporate a cross-attention texture decoder to secure the global style-consistency for each instance, producing more visually appealing and style-consistent textures for 3D-FRONT scenes compared against previous synthesis methods.

\end{itemize}

\section{Related work}

\mypara{Feed-forward 3D Generation.}
The advancement of 3D generation has adhered to the progress of 2D generative models. 
Adopting different backbone techniques, from variational autoencoders~\cite{vae}, generative adversarial networks~\cite{GANs}, autoregressive transformers~\cite{Transformer}, to the recent diffusion models~\cite{ho2020denoising,dhariwal2021diffusion}, 3D models have been trained on 3D data of various representations, including voxels~\cite{lin2023infinicity,siarohin2023unsupervised,smith2017improved,xie2018learning,chen2019learningnips}, point clouds~\cite{achlioptas2018learning,luo2021diffusion,pvd}, meshes~\cite{zhang2021sketch2model,pavllo2020convolutional}, signed distance functions~\cite{chen2019learning,cheng2022sdfusion,cheng2022cross,dai2021spsg,autosdf2022}, and more.
However, unlike the ubiquity of 2D images and videos, 3D data is inherently scarce and poses significant challenges in terms of acquisition and annotation.
Recent efforts have sought to address this issue by utilizing differentiable rendering techniques to learn from 2D images~\cite{EG3D, GRAF, StyleNeRF, Giraffe,piGAN, gao2022get3d,siddiqui2022texturify,yu2021learning, muller2023diffrf,skorokhodov3d,UVA,xu2022discoscene,abdal20233davatargan}. 
Although these models typically demonstrate proficiency in specific shape categories, they are incapable of handling 3D generation from free-form texts.

\begin{figure*}[!ht]
    \centering
    \includegraphics[width=\linewidth]{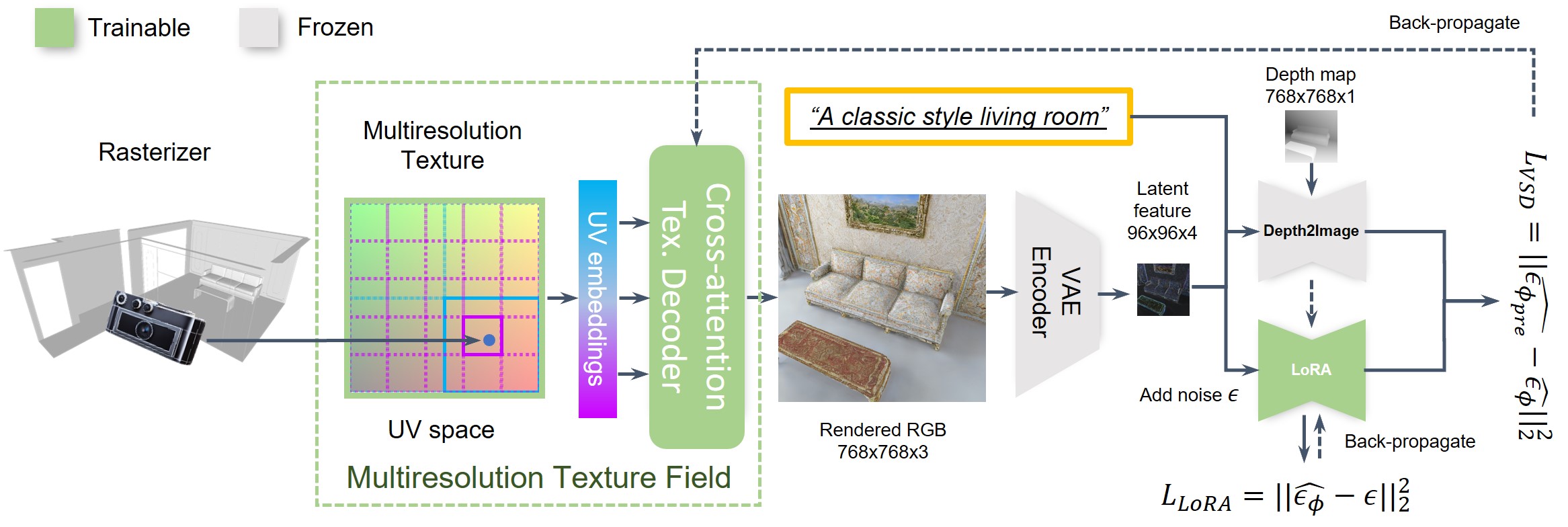}
    \caption{ 
    \textbf{Texture synthesis pipeline.} The target mesh is first projected to a given viewpoint via a rasterizer~\cite{liu2019soft}. Then, we render an RGB image with the proposed multiresolution texture field module. Specifically, each rasterized UV coordinate is taken as input to sample the UV embeddings from a multiresoultion texture. Afterward, the UV embeddings are mapped to an RGB image of shape $768 \times 768 \times 3$ via a cross-attention texture decoder. We use a pre-trained VAE encoder to compress the input RGB image to a $96 \times 96 \times 4$ latent feature. Finally, the Variational Score Distillation loss~\cite{wang2023prolificdreamer} is computed from the latent feature to update the texture field.
    }
    \label{fig:architecture}
\end{figure*}

\mypara{3D Generation with 2D diffusion models.}
Recently, significant strides have been made in the field of vision-language integration~\cite{hu2021unit, singh2022flava, wang2022ofa, radford2021learning, chen2022d3net, chen2020scanrefer, chen2021scan2cap, chen2022unit3d}
The advancements in text-to-image generation, particularly diffusion models trained on large-scale image collections \cite{dhariwal2021diffusion,ho2020denoising,ho2021cascaded,nichol2021improved,saharia2022image}, have prompted the integration of pretrained 2D diffusion models as priors to facilitate 3D generation. 
Two main streams of work have emerged.
The first branch directly incorporates the output of the 2D diffusion models along with the depth information. 
TEXTure~\cite{richardson2023texture} and Text2Tex~\cite{chen2023text2tex} perform texturing on given meshes with a depth-aware variation of diffusion models.
Other methods generate 3D scenes, where the geometry information is either jointly predicted~\cite{stan2023ldm3d} or obtained from off-the-shelf depth estimator~\cite{hollein2023text2room}.
The second branch of methods~\cite{poole2022dreamfusion,Fantasia3D,wang2023score,metzer2022latent,lin2022magic3d,wang2023prolificdreamer} attempt to distill knowledge from pretrained 2D diffusion models with the Score Distillation Sampling (SDS)~\cite{poole2022dreamfusion} technique and its subsequent improved variations in a per-prompt optimization manner. 
In contrast, we take advantages of the distilled knowledge from the depth-conditioned 2D diffusion priors to enable high-quality 3D texture synthesis.

\mypara{3D Scene Texturing.}
In this work, we focus on generating high-quality textures for 3D scenes. 
3D scene texturing has been studied by applying the 2D style transfer techniques~\cite{gatys2016image,gatys2017controlling,johnson2016perceptual} to 3D domain~\cite{chen2022upst,chiang2022stylizing,huang2022stylizednerf,zhang2022arf,haque2023instruct,hollein2022stylemesh}. 
However, these methods often emphasize low-level styles without semantic understanding. 
While existing 3D generation methods leveraging 2D diffusion models can theoretically be applied to 3D scene texturing, those based on inpainting \cite{richardson2023texture,chen2023text2tex} suffer from visible seams and accumulated artifacts, while distillation-based methods \cite{metzer2022latent} often produce blurry textures with incorrect geometry details.
In contrast, we optimize the target scene texture with accurate geometric cues and decode the high-resolution scene appearance via the proposed multiresolution texture field module, facilitating 3D scene texture synthesis with much better visual quality.

\section{Method}

The objective of our work is to texture an entire 3D scene with diffusion priors as the critic. 
In this section, we begin by introducing a Multiresolution Texture Field module to produce high-quality RGB textures, which consists of two key components: Multiresolution Texture and Cross-attention Texture Decoder.
The Multiresolution Texture is integrated to faithfully represent both the low- and high-frequency texture details at various scales (Sec.~\ref{sec:texture_field}).
Subsequently, to tackle the style-inconsistency issue brought by limited field of view and self-occlusion, the Cross-attention Texture Decoder module is incorporated to enforce a global style-awareness for each object in the scene (Sec.~\ref{sec:cross_attention}).
Finally, we adopt a pretrained diffusion model as training critic to dynamically distillate realistic scene appearance from the 2D depth-conditioned diffusion priors.  (Sec.~\ref{sec:vsd}).
The entire synthesis architecture is presented in Fig.~\ref{fig:architecture}.

\begin{figure}[!t]
    \centering
    \includegraphics[width=\linewidth]{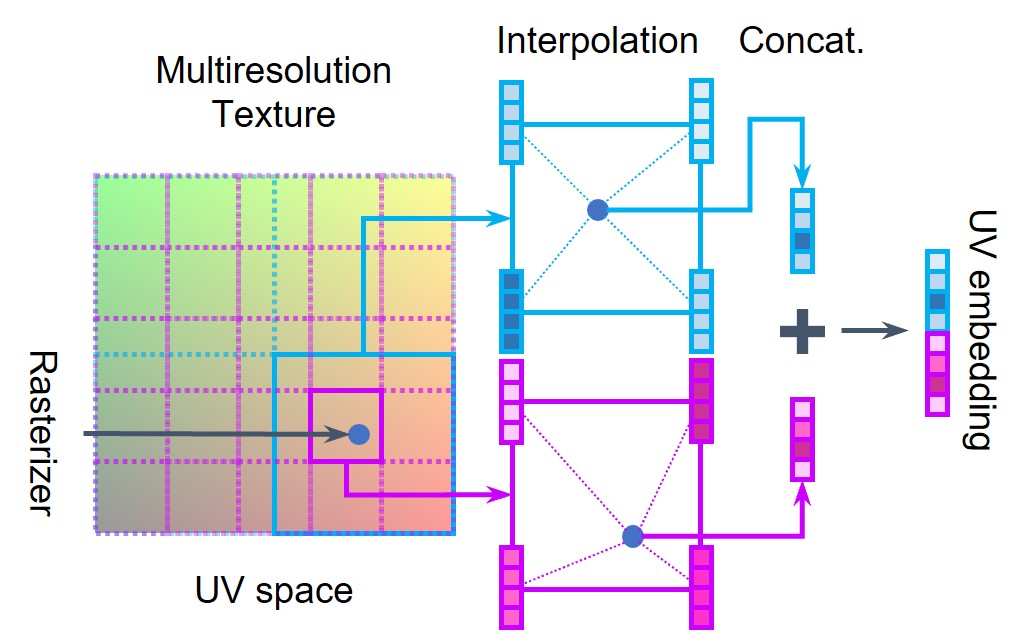}
    \caption{ 
    \textbf{Multiresolution Texture.}
    We use a multiresolution feature grid to encode positional features at different scale in the UV space. For a query UV coordinate, we interpolate the grid features at respective resolutions. The interpolated grid features are concatenated as the final UV embedding for the query UV coordinate.
    }
    \label{fig:hashgrid}
\end{figure}

\subsection{Multiresolution Texture Field}
\label{sec:texture_field}

The core of texture synthesis with 2D priors lies in generating RGB values visible to a series of pre-defined viewpoints. Previous methods maintain a $64 \times 64 \times 4$ latent map and operate directly on it with the SDS loss~\cite{poole2022dreamfusion, metzer2022latent}. This latent map is decoded via the variational autoencoder of the pre-trained diffusion model after convergence. The optimization process is technically view-consistent, as the diffusion priors are leveraged from numerous perspectives. Notwithstanding, we observe that the decoded RGB textures often carry patch-like artifacts and are subsequently inconsistent with the given geometry. This is caused by the mismatches between the low-resolution latent map and high-resolution RGB images, and the lack of perspective transformation of the same latent code in different views.

To tackle those inherent disadvantages of representing the target texture via a low-resolution latent map, we adopt an implicit texture field that queries the texture features with given UV coordinates. 
At its core, we integrate a multiresolution texture to prevent oversimplified appearance without any texture details. 
In particular, as shown in Fig.~\ref{fig:hashgrid}, we encode texture features for all query locations $q$ at each scale, and concatenate those features as the output UV embeddings $\mathcal{E}(q)$ to faithfully represent all texture details.
The UV embeddings are then decoded to the final RGB texture by the cross-attention texture decoder introduced in the next section.




\begin{figure}[!t]
    \centering
    \includegraphics[width=\linewidth]{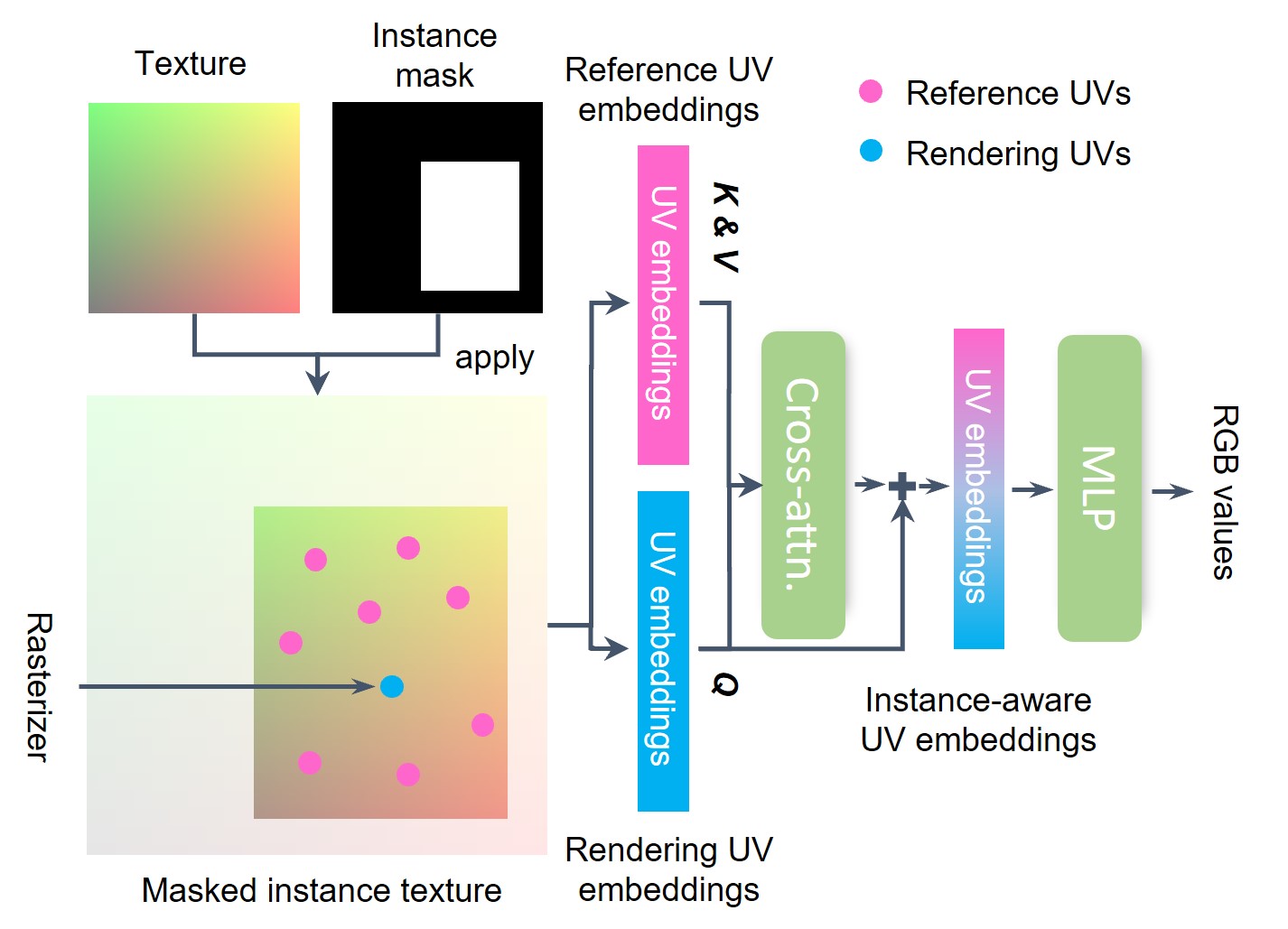}
    \caption{ 
    \textbf{Cross-attention Texture Decoder.} For each rasterized UV coordinate, we apply a UV instance mask to mask out the corresponding instance texture features. Then, we obtain the rendering UV embeddings for the rasterized locations in the view. At the same time, we extract the texture features for the pre-sampled UVs scattered across this instance as the reference UV embeddings. We deploy a multi-head cross-attention module to produce the instance-aware UV embeddings. Here, we treat the rendering UV embeddings as the Query, and the reference UV embeddings as the Key and Value. Finally, a shared MLP maps the instance-aware UV embeddings to RGB values in the rendered view. 
    }
    \label{fig:cross_attention}
\end{figure}

\subsection{Cross-attention Texture Decoder}
\label{sec:cross_attention}

Since the texture is optimized in image space, instance textures are often constrained by limited field of view and self-occlusion. As a result, the optimized texture often suffers from style-inconsistency. Therefore, we propose a simple yet effective rendering module with global instance awareness to predict RGB values from UV embeddings. This is done by incorporating a multi-head cross-attention module to the texture features. As Fig~\ref{fig:cross_attention} illustrates, for each rasterized UV coordinate, we apply a UV instance mask to mask out the corresponding instance texture features. Then, we obtain the rendering UV embeddings for the rasterized locations in the view. At the same time, we extract the texture features for the pre-sampled UVs scattered across this instance as the reference UV embeddings. We deploy a multi-head cross-attention module to produce the instance-aware UV embeddings. Here, we treat the rendering UV embeddings as the Query, and the reference UV embeddings as the Key and Value. Finally, a shared MLP maps the global-aware UV embeddings to RGB values in the rendered view. We denote the whole rendering process as $\mathcal{C}=f(\mathcal{E}(q); \theta)$, where $\mathcal{C}$ represents an RGB image at arbitrary resolution, $f(\theta)$ is a differentiable function resembles the entire texture field with trainable parameters $\theta$.

\subsection{Texture Field Optimization via VSD}
\label{sec:vsd}

We adopt a pre-trained ControlNet model as a critic to optimize the texturing module $f(\theta)$ following the strategy of Latent-Paint~\cite{metzer2022latent}, as shown in Fig.~\ref{fig:architecture}. Here, the UNet of a pre-trained latent diffusion model (LDM)~\cite{Rombach_2022_CVPR} applied to calculate the gradients based on a low-resolution $96 \times 96$ latent map. We observe that such low-resolution rendering often lead to broken visual quality and unsatisfactory view consistency. This is primarily due to the size mismatches between the $96 \times 96$ optimization target and the final $768 \times 768$ RGB output. Additionally, prior work exclude geometric cues from the diffusion priors, resulting in poor consistency between the generated textures and target geometry. To address those issues, we directly render an $768 \times 768$ RGB image via querying the texture field $\mathcal{C}=f(\mathcal{E}(q); \theta)$. In each iteration, we first optimize $\mathcal{C}$ via the VSD objective~\cite{wang2023prolificdreamer} with a pre-trained frozen depth-conditioned diffusion prior $\phi_{\text{pre}}$ and a trainable LoRA module $\phi$:
\begin{equation}
\label{eq:vsd}
    \mathcal{L}_{\text{VDS}}(\theta) := \mathbb{E}_{t, \epsilon}[w(t)(\epsilon_{\phi_{\text{pre}}}-\epsilon_{\phi})\frac{\partial f(\theta)}{\partial \theta}]
\end{equation}
where $\epsilon_{\phi_{\text{pre}}}=\phi_{\text{pre}}(f(\theta);y,d,t)$ and $\epsilon_{\phi}=\phi(f(\theta);y,d,t))$. We draw time step randomly by $t \sim \mathcal{U}(0.02, 0.98)$. The injected noise is $\epsilon \sim \mathcal{N}(0, 1)$. $d$ is the depth map in the current viewpoint produced by the rasterizer. $y$ is the noised input to the UNet. The weighting function $w(t)$ is empirically set as $w(t)=\sqrt{1-\prod_{s=1}^t\alpha_s}$. Note that $\phi_{\text{pre}}$ and $\phi$ are kept frozen when updating the parameters of the texture field $\theta$. After $\mathcal{C}=f(\mathcal{E}(q); \theta)$ is updated via the gradients of VSD, we unfreeze and update the LoRA module $\phi$:
\begin{equation}
    \mathcal{L}_{\text{LoRA}}(\phi) = \min_{\phi}\sum_{i=1}^{n}\mathbb{E}_{t, \epsilon}[||\epsilon_{\phi}(f(\theta);y,d,t)-\epsilon||_2^2]
\end{equation}

\subsection{Inference}

Since the texture field $f(\theta)$ only receives the UV coordinates as input, producing the final RGB texture is straightforward. For each position $q_i$ in UV space, the corresponding pixel value $c_i$ of the output RGB texture $\mathcal{C}$ can be simply queried by $c_i=f(\mathcal{E}(q_i); \theta)$. Thanks to the multiresolution grid encoding, it is worth mentioning that there is no specification for the size of the final RGB texture, i.e. the resolution of the texture can be adjusted according to the computational resources.

\section{Results}

\begin{figure*}[!ht]
    \centering
    \includegraphics[width=\linewidth]{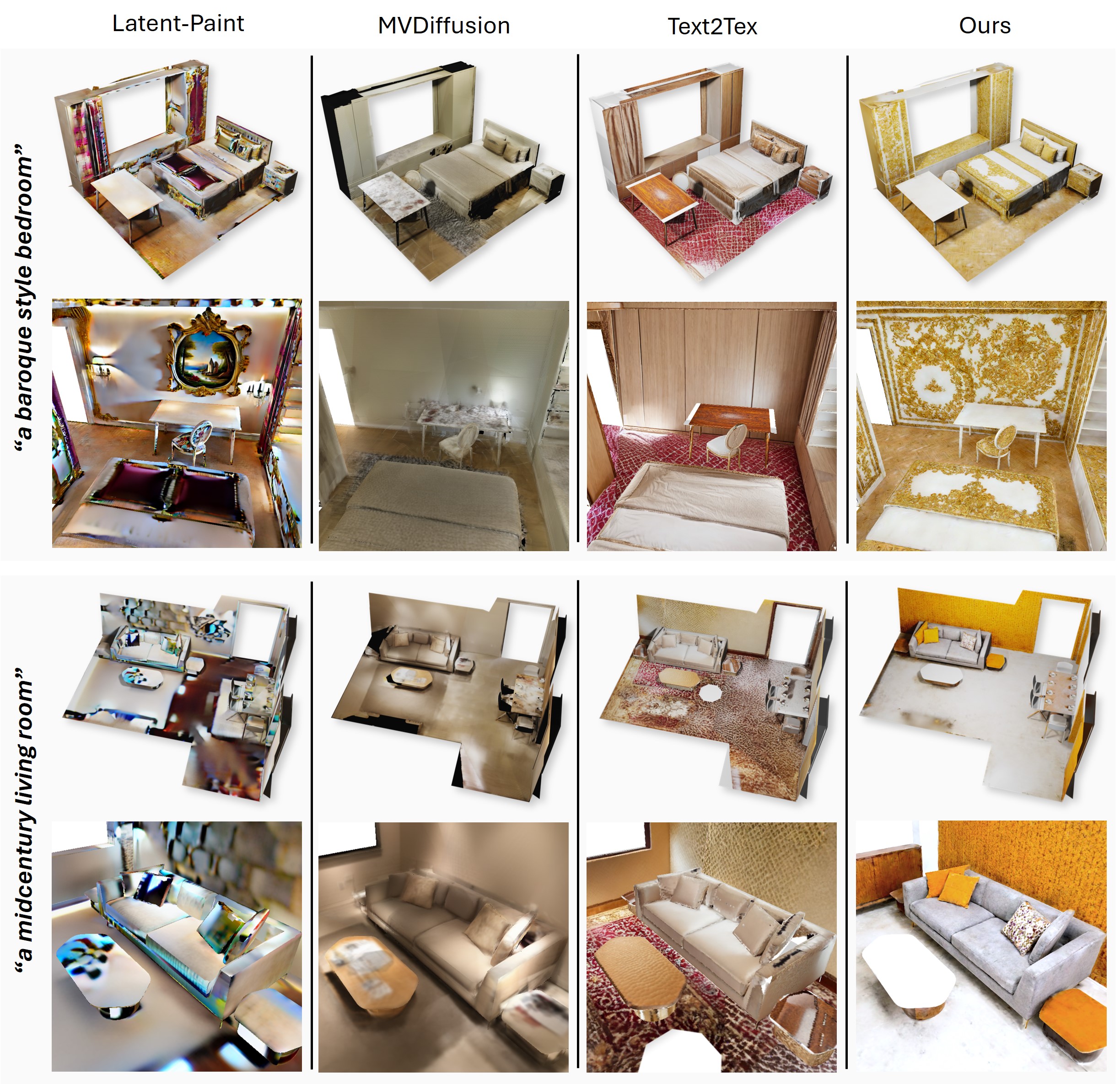}
    \caption{ 
    \textbf{Qualitative comparisons. }
    Latent-Paint~\cite{metzer2022latent} suffers from over-saturation and hallucinates scene components. MVDiffusion~\cite{tang2023mvdiffusion} delivers blurry textures and fails to reflect the input prompts. Text2Tex~\cite{chen2023text2tex} struggles to keep all instances style-consistent. In contrast, our method produces high-quality textures and maintains overall style-consistency across instances in the scenes. Ceilings and back-facing walls are excluded for better visualizations. Images best viewed in color.
    }
    \label{fig:baselines}
\end{figure*}

\begin{figure*}[!ht]
    \centering
    \includegraphics[width=0.92\linewidth]{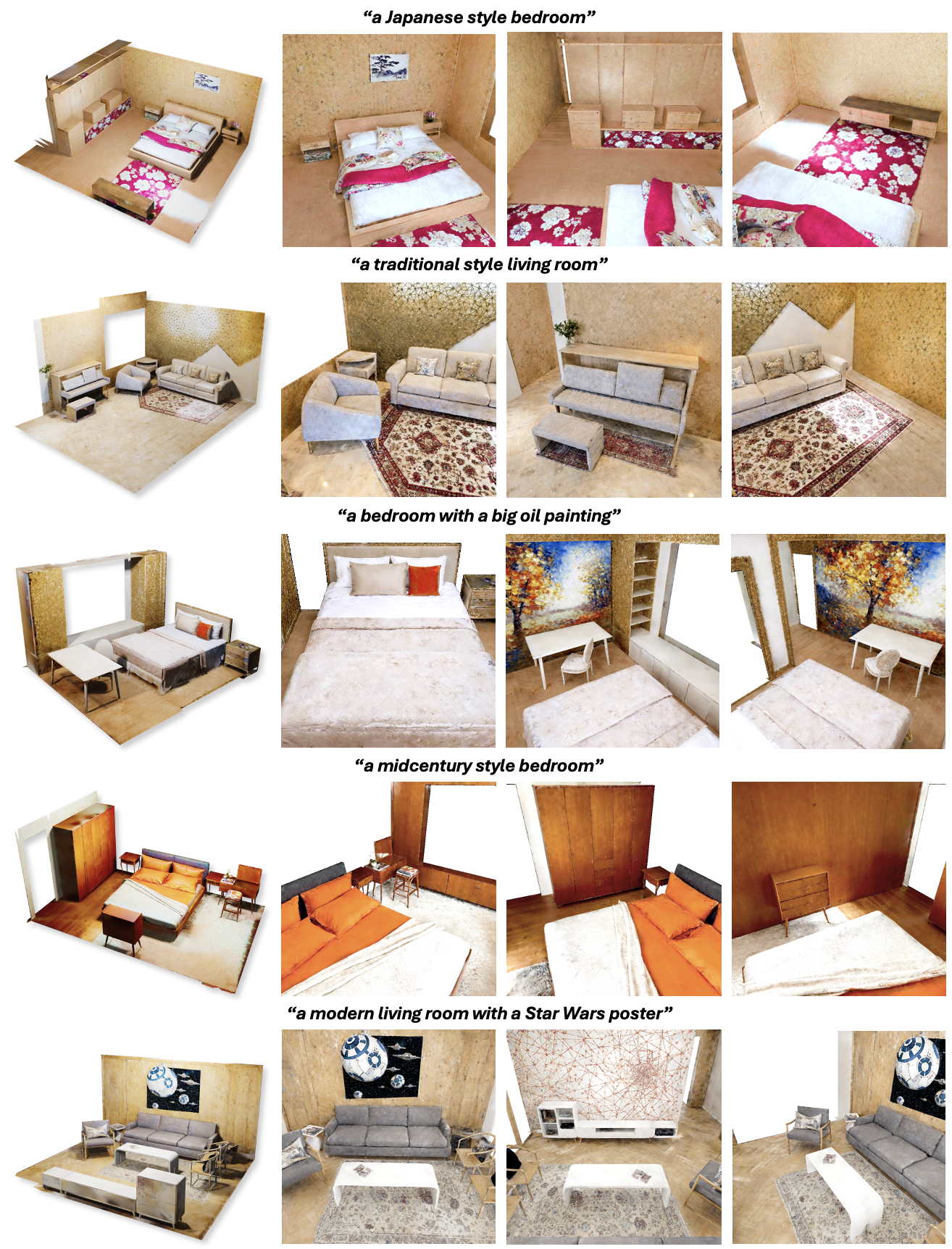}
    \caption{ 
    \textbf{Synthesized textures for 3D-FRONT scenes.}
    Our method generates high-quality style-coherent textures, and reflects the iconic traits in the prompts. 
    Ceilings and back-facing walls are excluded for better visualizations. Images best viewed in color.
    }
    \label{fig:top_and_closeup}
\end{figure*}

\begin{figure*}[!ht]
    \centering
    \includegraphics[width=\linewidth]{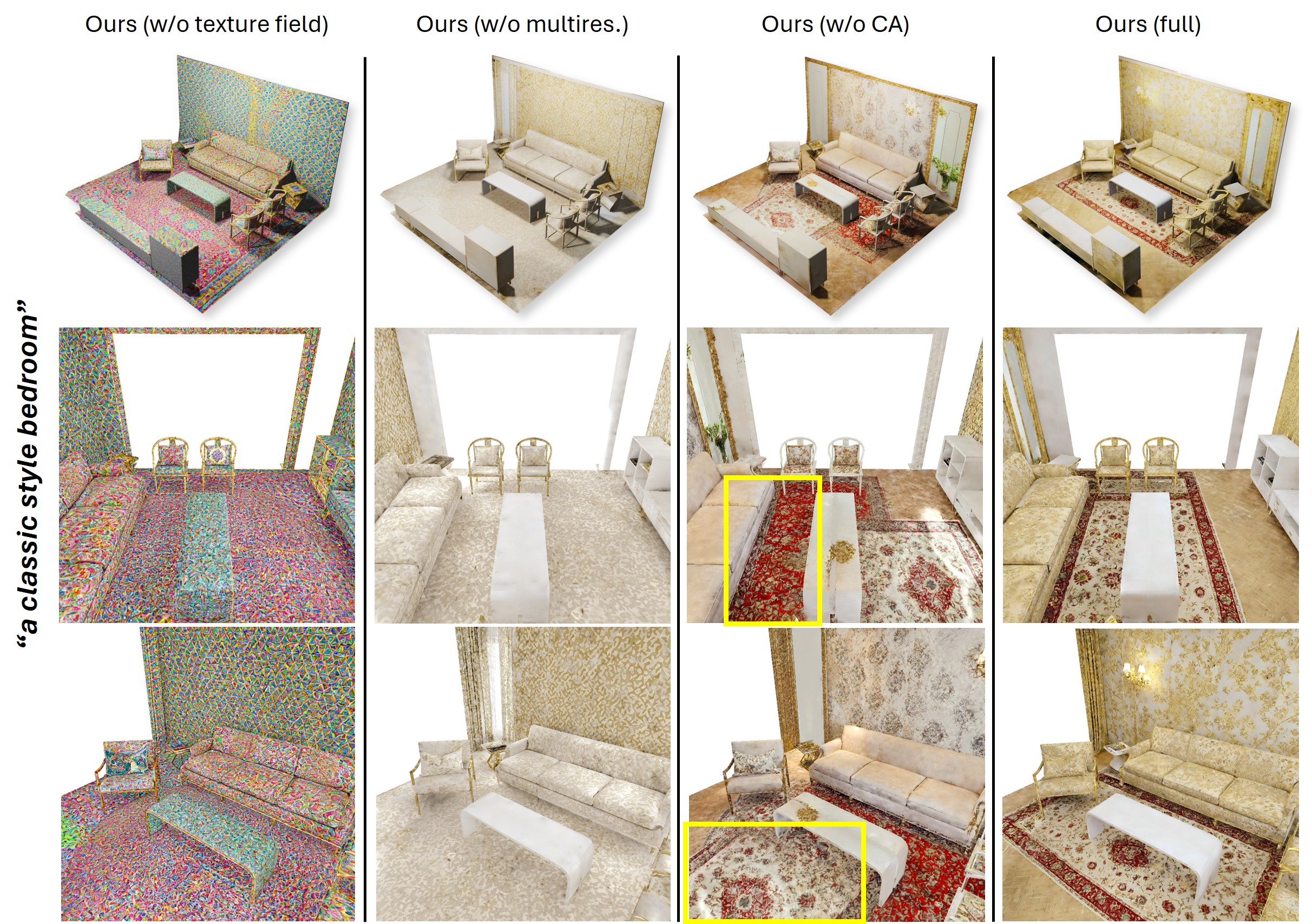}
    \caption{ 
    \textbf{Ablation studies on the key components.}
    Optimizing an RGB texture directly without the proposed texture field results in extreme noisy and unrealistic textures. A single-resolution texture fails to capture texture details and produces bubble artifacts. Removing the cross-attention decoder leads to style inconsistency, especially for big instances such carpet, as shown in the yellow boxes. In contrast, our full method produces high-quality and style-consistent textures without aforementioned artifacts.
    }
    \label{fig:ablations}
    \vspace{-3mm}
\end{figure*}

\subsection{Implementation Details}

We apply the ControlNet Depth model~\cite{zhang2023adding} for VSD optimization. 
The multiresolution texture is implemented by multiresolution hash encoding~\cite{muller2022instant}.
During each optimization iteration, we randomly pick 1 viewpoint from all perspective points scattered across the scene. 
We set the learning as $0.001$ for optimizing the texture field and $0.0001$ for fine-tuning the LoRA module.
The entire optimization uses $5,000$ viewpoints and takes $30,000$ iterations to converge.
To generate a more visually appealing appearance via VSD, We adopt the time annealing scheme following ProlificDreamer~\cite{wang2023prolificdreamer}, where we sample time steps $t \sim \mathcal{U}(0.02, 0.98)$ for the first $5,000$ steps and then anneal into $t \sim \mathcal{U}(0.02, 0.50)$ for the rest of the optimization.
For the proposed cross-attention decoder, we pre-sampled $4,096$ UV coordinates scattered across each instance.
To enable cross-attention in such a long context, we implement the cross-attention module with Flash Attention v2~\cite{dao2023flashattention}.
Each synthesis process takes around 20 hours to converge on an NVIDIA RTX A6000. 
After convergence, we generate a high-resolution $4,096 \times 4,096$ RGB image as the final scene texture.
Our implementation uses the PyTorch~\cite{paszke2017automatic} framework, with PyTorch3D~\cite{ravi2020pytorch3d} for rendering and texture projection.













\subsection{Quantitative Analysis}

We compare our method against texture synthesis methods appeared recently in Tab.~\ref{tab:quantitatives}, including Latent-Paint~\cite{metzer2022latent}, MVDiffusion~\cite{tang2023mvdiffusion}, and Text2Tex~\cite{chen2023text2tex}. We experiment all methods on 10 3D-FRONT~\cite{fu20213dfront} scenes with 2 different text prompts for each scene. Here, we calculate CLIP score (CLIP)~\cite{radford2021learning} and Inception Score (IS)~\cite{smith2017improved} to measure the fidelity with input prompts and texture quality, respectively. Our method outperforms all baselines on the 2D automated metrics by a significant margin. We additionally report the User Study results from 75 participants about the Visual Quality (VQ) and Prompt Fidelity (PF) on a scale of 1-5. 
Our method is shown to be more favored by human users.

\begin{table}[!t]
    \centering
    \resizebox{\linewidth}{!}{
        \begin{tabular}{l cc cc}
            \toprule
                \multirow{2}{*}{Method} & \multicolumn{2}{c}{2D Metrics} & \multicolumn{2}{c}{User Study} \\
                                \cmidrule(l{2pt}r{2pt}){2-3} \cmidrule(l{2pt}r{2pt}){4-5}
            & CLIP $\uparrow$ & IS $\uparrow$ & VQ $\uparrow$ & PF $\uparrow$\\
            \midrule
            Latent-Paint~\cite{metzer2022latent} & 18.37 & 1.96 & 1.57 & 2.11 \\
            MVDiffusion~\cite{tang2023mvdiffusion} & 18.47 & 2.83 & 3.09 & 3.12 \\
            Text2Tex~\cite{chen2023text2tex} & 20.83 & 2.87 & 2.62 & 3.04 \\
            \midrule
            \ARCH (Ours) w/o texture field & 15.77 & 1.56 & 1.23 & 1.11 \\
            \ARCH (Ours) w/o multires. tex. & 19.87 & 2.79 & 2.11 & 2.39 \\
            \ARCH (Ours) w/ cross-attn. & 20.94 & 3.29 & 3.94 & 4.05 \\
            \ARCH (Ours) & \textbf{22.18} & \textbf{3.33} & \textbf{4.40} & \textbf{4.29} \\
            \bottomrule
        \end{tabular}
    }
    \caption{
    \textbf{Quantitative comparisons.}
    We report the 2D metrics and User Study results for quantitative comparisons, including: CLIP score (CLIP)~\cite{radford2021learning}, Inception Score (IS)~\cite{smith2017improved}, Visual Quality (Visual Quality), and Prompt Fidelity (PF). We show that our method produces textures with the highest quality.
    }
    \label{tab:quantitatives}
\end{table}

\subsection{Qualitative Results}

We show the qualitative comparisons in Fig.~\ref{fig:baselines}. Latent-Paint suffers from the over-saturation issue and hallucinates non-existing objects, such as the huge frame on the wall (see the first example in the first row in Fig.~\ref{fig:baselines}). Those unrealistic texture components are produced by the inaccurate geometric cues and the mismatch between the optimized latent representation and final texture. MVDiffusion~\cite{tang2023mvdiffusion} produces overall smooth but blurry and dimmed texture. It also fails to reflect the iconic properties in the prompts, such as ``baroque'' and ``luxury''. Text2Tex~\cite{chen2023text2tex} generates plausible textures for individual objects, but fails to achieve global style consistency across objects. In contrast, our method synthesizes high-quality with overall coherent styles within and across objects, reflecting the representative traits in the prompts with high-fidelity (see the baroque paintings above and the golden pillows below). We additionally visualize our texture synthesis results for different 3D-FRONT~\cite{fu20213dfront} scenes and input prompts in top-down views and close-up renderings in Fig.~\ref{fig:top_and_closeup}, further demonstrating the supreme texture quality and fidelity produced by our method.

\subsection{Ablation Studies}

We conduct ablation experiments on the key components of our method, including multiresolution texture (Sec.~\ref{sec:texture_field}), and cross-attention decoder (Sec.~\ref{sec:cross_attention}). All comparisons are shown in Tab.~\ref{tab:quantitatives} and Fig.~\ref{fig:ablations}.

\paragraph{Does texture field produce better textures than RGB tensors?}

Since only a few UVs are sampled by the rasterizer during each iteration, directly optimizing an RGB tensor as the output texture leads to noisy artifacts, as shown in the first column in Fig.~\ref{fig:ablations}. Additionally, the optimization is often difficult to converge with different gradient scales across the whole RGB tensor. As a result, the optimized texture appears to be broken and unrealistic. In contrast, the MLP in the proposed texture field module effectively smoothens the back-propagated gradients, producing much smoother and detailed textures.

\paragraph{Does multiresolution texture improve the visual quality?}

As previous studies indicate, implicit representations via MLPs tend to learn low-frequency information~\cite{sitzmann2020implicit, tancik2020fourier}. As such, the texture field with single resolution produces an over-simplified appearance without texture details. Such texture lacks the characteristic properties of the input prompt, and carries noisy bubble artifacts, as shown in the second column in Fig.~\ref{fig:ablations}. We show that the multiresolution texture is capable of producing a visually appealing and highly detailed mesh appearance.

\paragraph{Does cross-attention strengthen the style consistency?}

Replacing the cross-attention decoder module with a simple MLP also produces plausible textures. However, such replacement exposes global style inconsistency issue. Due to limited field of view and self-occlusion, the appearance of the same object can be synthesized differently. As shown in Fig.~\ref{fig:ablations}, big objects such as the carpet do not share a coherent pattern. It is difficult for the big objects to maintain style consistency during optimization, if there is no global information shared across views. The proposed cross-attention decoder effectively tackles this issue by globally sharing the style features within each object. This enforces the instance style awareness, and therefore produces more style-consistent textures for all instances across the scene.

\subsection{Limitations}

Although our method enables high-quality texture synthesis for indoor scenes, we still notice that our method tends to generate textures with shading effects. This phenomenon becomes more obvious when the scene structure indicates the existence of lighting such as lamp, window, or even mirror. We believe this issue can be properly addressed by carefully fine-tuning the diffusion priors on the indoor scene images without shading effects. We acknowledge this challenge and leave it to future research.

\section{Conclusion}

We introduce SceneTex, a novel method for effectively generating high-quality and style-consistent textures for indoor scenes using depth-to-image diffusion priors. 
At its core, SceneTex proposes a multiresolution texture field to implicitly encode the mesh appearance. 
We optimize the target texture via a score-distillation-based objective function in respective RGB renderings. 
To further secure the style consistency across views, we introduce a cross-attention decoder to predict the RGB values by cross-attending to the pre-sampled UV coordinates within each instance.
We show that the proposed texture field with multiresolution texture is capable of generating visually appealing high-quality texture.
Moreover, the proposed cross-attention decoder further strengthens the global style awareness for each instance, resulting in style-coherent appearance in the target scene.
Extensive analysis show that SceneTex enables various and accurate texture synthesis for 3D-FRONT scenes, demonstrating significant improvements in visual quality and prompt fidelity over the prior texture generation methods.
Overall, we hope our work can inspire more future research in the area of text-to-3D generation.
\section*{Acknowledgements}
This work was supported by a gift by Snap Inc., the ERC Starting Grant Scan2CAD (804724), and the German Research Foundation (DFG)
Research Unit “Learning and Simulation in Visual Computing”.
We thank Angela Dai for the video voiceover.


{
    \small
    \bibliographystyle{ieeenat_fullname}
    \bibliography{main}
}



\clearpage
\appendix

\section*{Supplementary Material}

In this supplementary material, we present additional results and analysis in Sec.~\ref{sec:additional_qualitative} to show the effectiveness of the proposed texture synthesis method.
We also provide details of the user study in Sec.~\ref{sec:user_study_detail}.
For reproducibility, we produce the architecture details of the proposed Cross-attention Texture Decoder in Sec.~\ref{sec:ca_detail}, and the hash encoding~\cite{muller2022instant} configurations of the multiresolution texture in Sec.~\ref{sec:hashgrid_detail}.

\section{Additional Qualitative Results}
\label{sec:additional_qualitative}

To further showcase the effectiveness of the proposed method, we present more texture synthesis results on 3D-FRONT dataset~\cite{fu20213dfront}.

\begin{figure*}[!ht]
    \centering
    \includegraphics[width=0.92\linewidth]{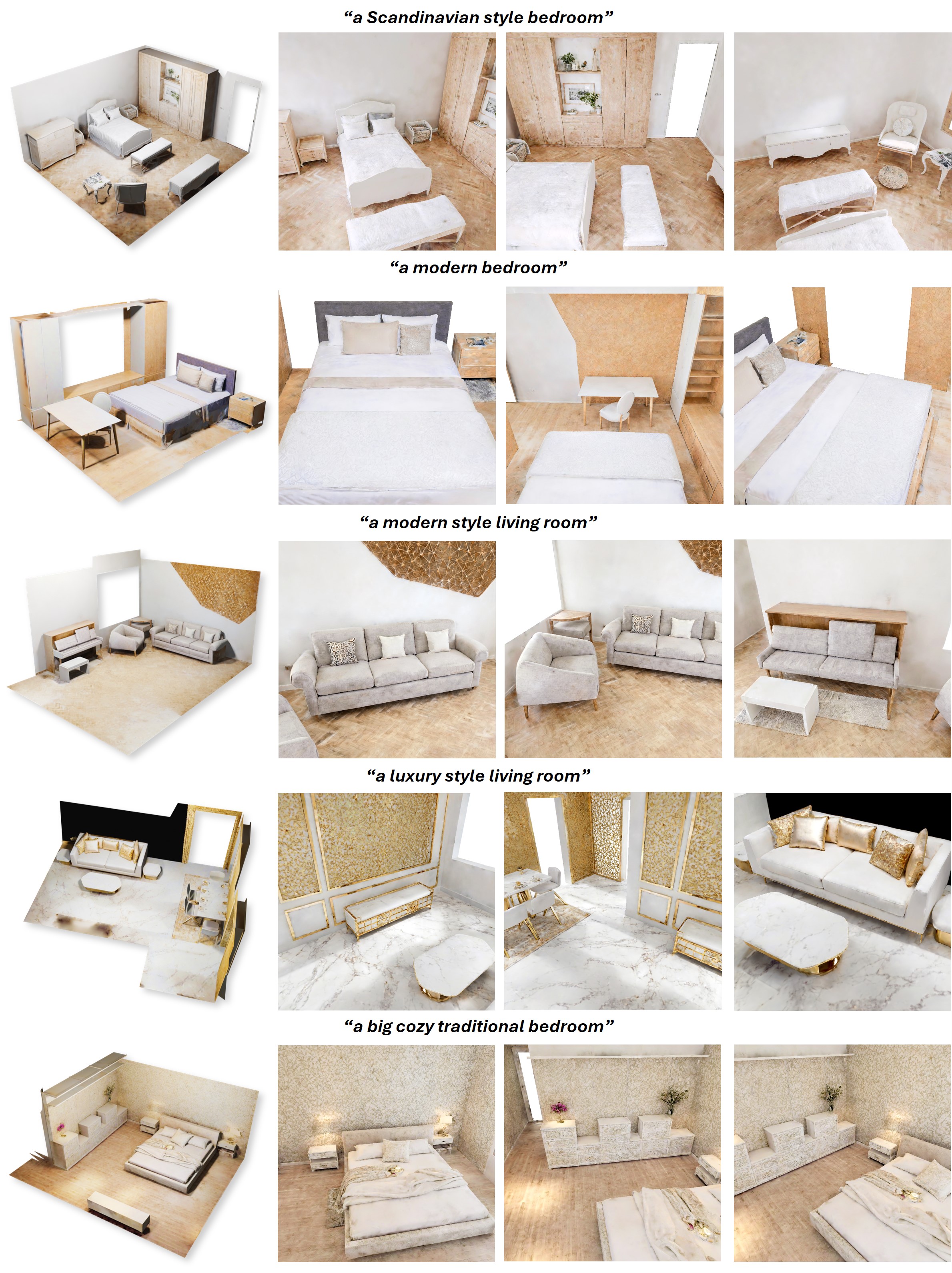}
    \vspace{-3mm}
    \caption{ 
    \textbf{Synthesized scene textures with modern styles.}
    Our method is capable of generating high-quality scene appearance for common indoor decoration styles. Ceilings and back-facing walls are excluded for better visualizations. Images best viewed in color. 
    }
    \label{fig:modern}
\end{figure*}

\begin{figure*}[!ht]
    \centering
    \includegraphics[width=0.92\linewidth]{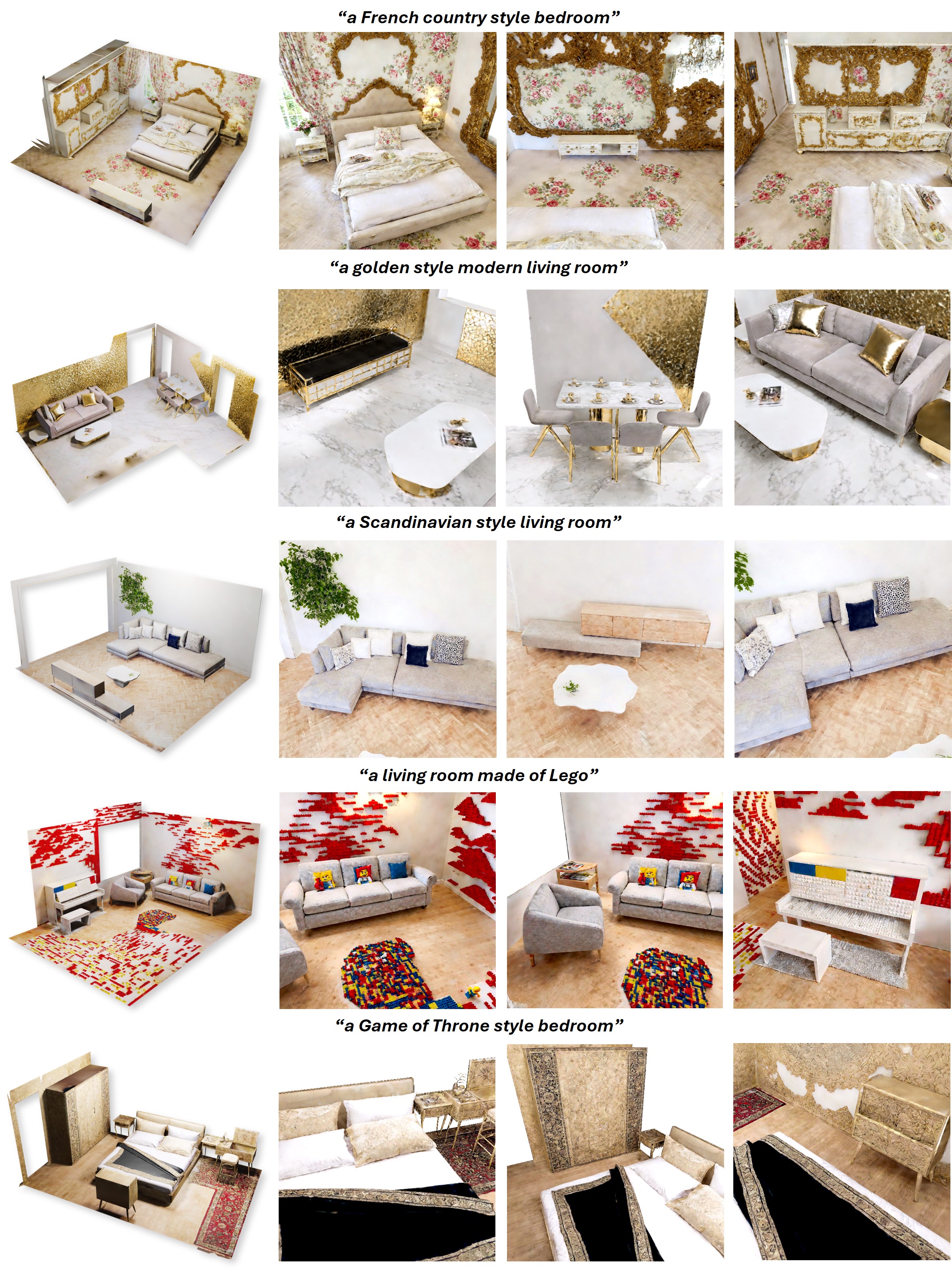}
    \vspace{-3mm}
    \caption{ 
    \textbf{Synthesized scene textures with creative styles.}
    Our method can also generate textures for challenging and creative scene styles. Ceilings and back-facing walls are excluded for better visualizations. Images best viewed in color. 
    }
    \label{fig:creative}
\end{figure*}

\paragraph{Realistic indoor scene decorations.}
Our method is capable of generating high-quality realistic scene appearance for most common indoor decoration styles, such as ``Scandinavian'' and ``luxury'', as shown in Fig.~\ref{fig:modern}.
Additionally, our method can move beyond common indoor styles to more challenging ones, such as ``Lego'' and ``Game of Throne'', as shown in Fig.~\ref{fig:creative}.
It demonstrates the flexibility of our method for synthesizing high-quality texture while loyally reflecting iconic properties in the input prompts.
We believe that those synthesis results indicate a great potential to stylize more high-quality 3D textures for indoor scene.

\begin{figure*}[!ht]
    \centering
    \includegraphics[width=\linewidth]{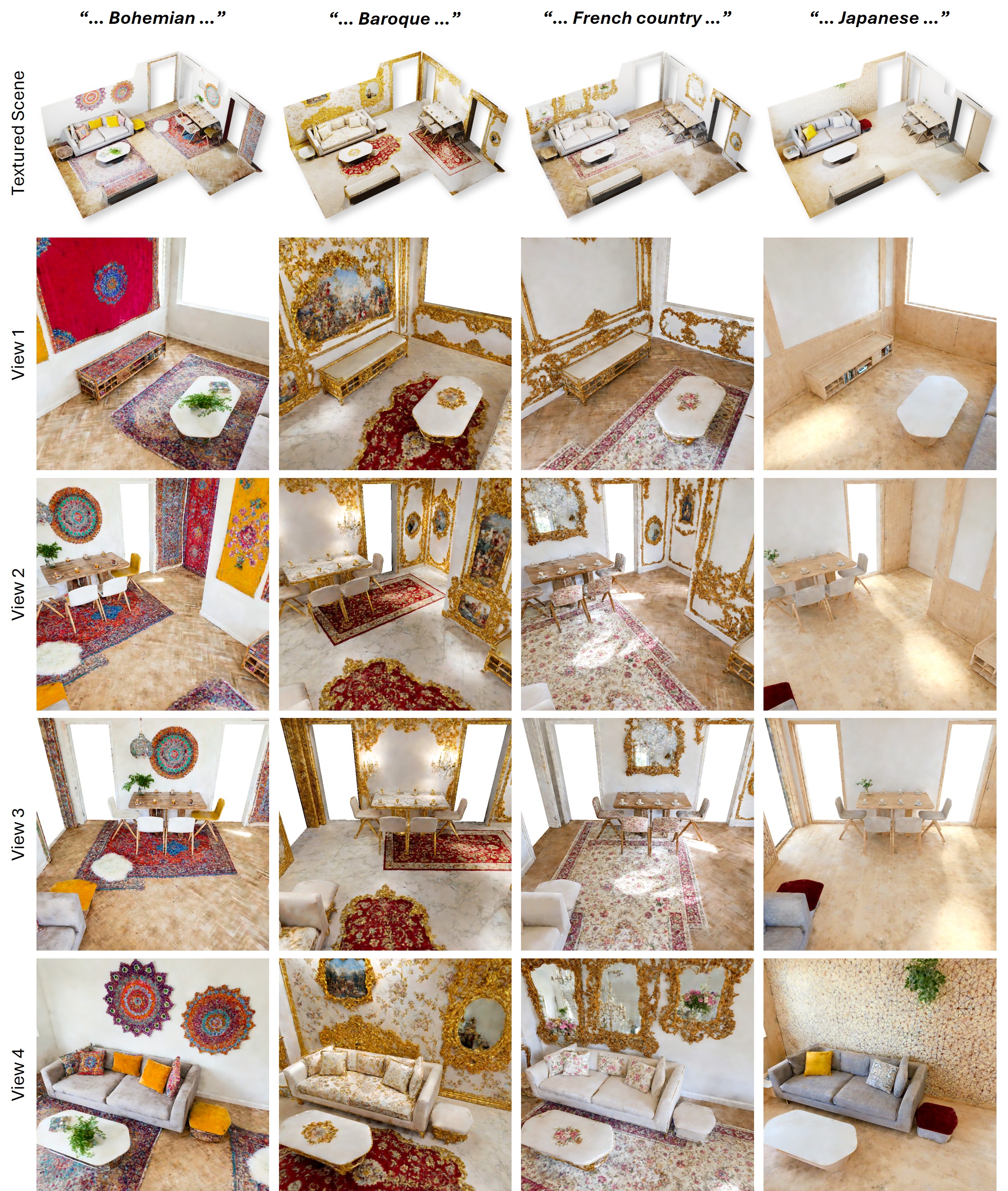}
    \caption{ 
    \textbf{Synthesized textures for 3D-FRONT scenes.}
    Our method generates different textures for the same input scene. We use the prompt template ``a \textlangle STYLE\textrangle~living room'' with 4 different styles for texture generation.
    }
    \label{fig:various_prompts}
\end{figure*}

\paragraph{Stylizing the same scene.}
In Fig.~\ref{fig:various_prompts}, we show that our method can generate different texturing results on the same objects. 
Here, we use the same prompt template ``a \textlangle STYLE\textrangle~living room'' with 4 different styles (``Bohemian'', ``Baroque'', ``French country'', and ``Japanese'').
All our textures are highly detailed and loyal to the style, demonstrating a great potential to generate diverse and various 3D scenes appearances. 

\section{User Study Details}
\label{sec:user_study_detail}

\begin{figure*}[!ht]
    \centering
    \includegraphics[width=\linewidth]{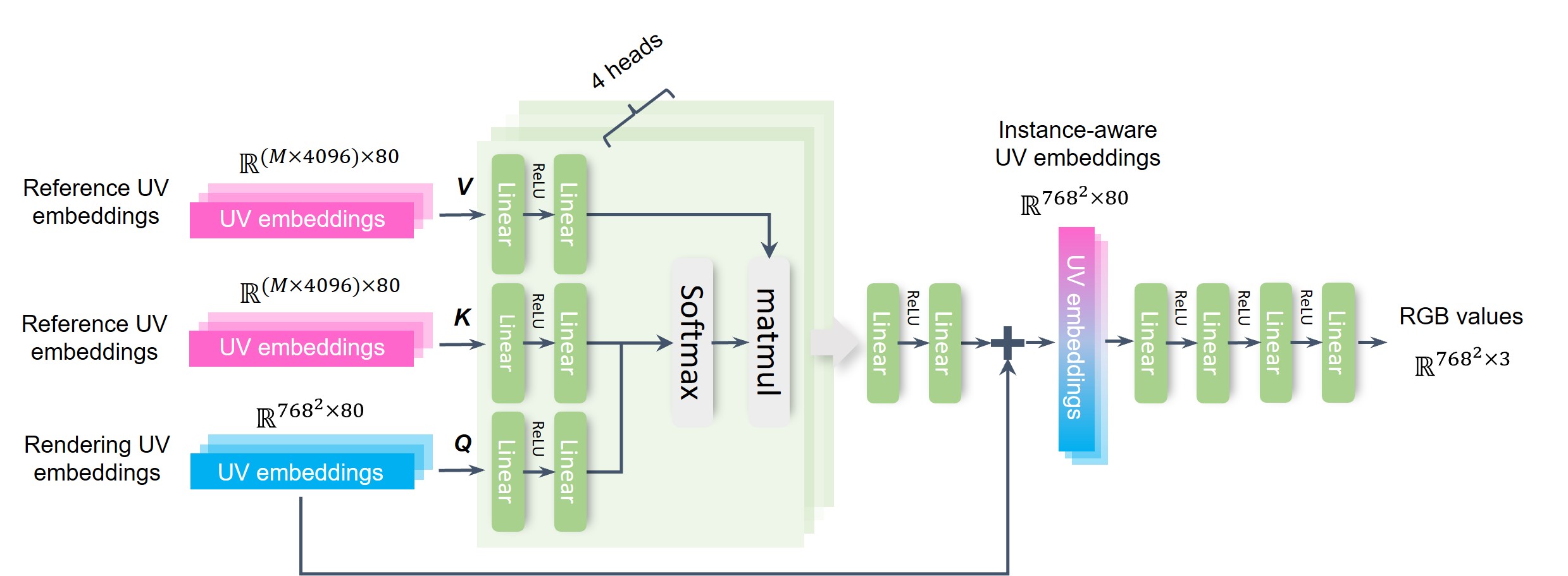}
    \caption{ 
    \textbf{Architecture details for Cross-attention Texture Decoder.} 
    For the target $768\times768$ image, we directly obtain the rendering UV embeddings in shape $\mathbb{R}^{768^2\times80}$.
    Then, for the current scene with $M$ objects, we query the reference UV embeddings in shape $\mathbb{R}^{(M\times4096)\times80}$ of the pre-sampled $4,096$ positions for each object.
    We apply 2-layer MLPs to map the UV embeddings to the Key, Value, and Query for the cross-attention module.
    We use a 4-head cross-attention module implemented with Flash Attention v2~\cite{dao2023flashattention}.
    The output features are mapped with a 2-layer MLP, and added to the original rendering UV embeddings as the final instance-aware UV embeddings.
    Finally, a 4-layer MLP predicts the RGB renderings of the current viewpoint.
    }
    \label{fig:cross_attention_detail}
\end{figure*}

\begin{figure*}[!ht]
    \centering
    \includegraphics[width=\linewidth]{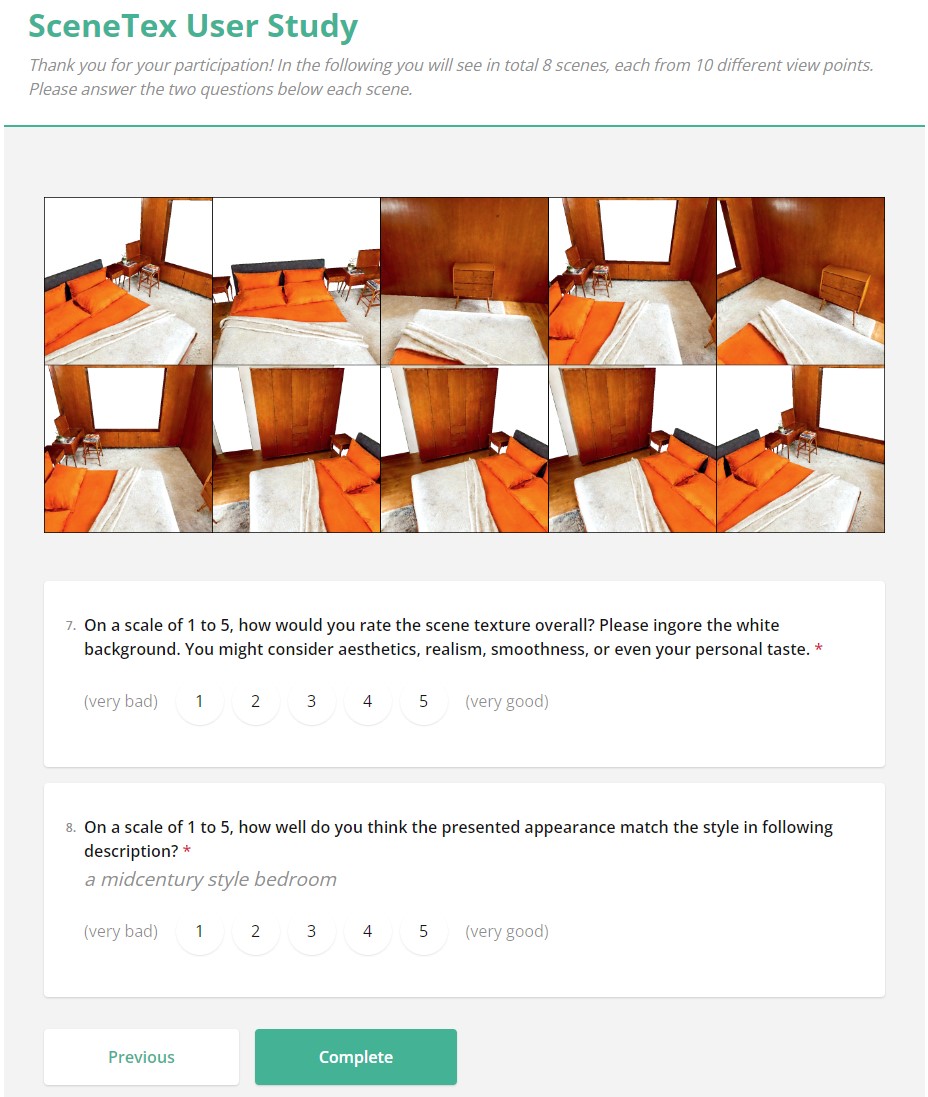}
    \caption{ 
    \textbf{Screenshot of the user study interface.}
    We present 10 rendered views from 8 different texturing results to each human user, and ask them to rate the appearance and the similarity with input prompts on a scale of 1-5.
    }
    \label{fig:user_study}
\end{figure*}

We develop a Django-based web application for the user study. In Fig.~\ref{fig:user_study}, we show the interface for the questionnaire. 
We randomly select textured scenes from each baseline and our method to form a batch of 8 samples. 
To better visualize the textured scene, we render multi-view images for those scenes from 10 different viewpoints.
After presenting the 10 rendered images, we ask the users to rate the appearance of the scenes from 1 to 5 in terms of aesthetics, realism, smoothness, etc.
We also ask the users to measure how well the scene appearances match the given descriptions.
To avoid biases and cheating in this user study, we shuffle scenes so that there is no positional hint of our method. 
In the end, we gather 100 responses from 75 participants to calculate the user study results.

\section{Cross-attention Texture Decoder}
\label{sec:ca_detail}

We implemented a neural texture decoder with cross-attention mechanism to strengthen the style-consistency within each object in the scene. As Fig.~\ref{fig:cross_attention_detail} shows, we map the UV embeddings with 2-layer MLPs in a 4-head cross-attention module implemented with Flash Attention v2~\cite{dao2023flashattention}. A skip connection is applied between the original rendering UV embeddings and the output features as the final instance-aware UV embeddings. The final RGB renderings are produced by a 4-layer MLP with hidden size 256.

\section{Multiresolution Texture Configuration}
\label{sec:hashgrid_detail}

\begin{table}[!t]
    \centering
        \begin{tabular}{l r}
            \toprule
            Parameter & Value\\
            \midrule
            Number of levels & 20 \\
            Hash table size & $2^{24}$ \\
            Number of feature dimensions & 4 \\
            Min. resolution & $16^2$ \\
            Max. resolution & $4096^2$ \\
            \bottomrule
        \end{tabular}
    \caption{
    \textbf{Multiresolution texture configuration.}
    We configure the hash encoding parameters following iNGP~\cite{muller2022instant}.
    }
    \label{tab:hashgrid}
\end{table}

In Tab.~\ref{tab:hashgrid}, we illustrate the implementation details of our multiresolution texture for reproducibility, following the configuration in iNGP~\cite{muller2022instant}.
Here, we use a deep hierarchy with 20 layers of texture features to learn scene appearance at various scales.
This ensures our synthesis architecture to faithfully encode detailed texture information irrespective of the distance between the viewpoint and mesh surface.
To reduce possible hash collision in UV space, we choose a comparatively big hash table size of $2^{24}$.
Our multiresolution texture encodes texture features from $16 \times 16$ to $4,096 \times 4,096$.

\end{document}